# A Session Based Blind Watermarking Technique within the NROI of Retinal Fundus Images for Authentication Using DWT, Spread Spectrum and Harris Corner Detection

Nilanjan Dey[1], Moumita Pal[2], Achintya Das[3]

[1]Asst. Professor Dept. of IT, JIS College of Engineering, Kalyani, WB. India.
[2]Asst. Professor Dept. of ECE, JIS College of Engineering, Kalyani, WB. India.
[3] Professor and Head, Elec. and Telecom Engg Dept., Kalyani Govt. Engg. College, Kalyani,WB,India.

**ABSTRACT**

**Digital Retinal Fundus Images** helps to detect various ophthalmic diseases by detecting morphological changes in optical cup, optical disc and macula. Present work proposes a method for the authentication of medical images based on Discrete Wavelet Transformation (DWT) and Spread Spectrum. Proper selection of the Non Region of Interest (NROI) for watermarking is crucial, as the area under concern has to be the least required portion conveying any medical information. Proposed method discusses both the selection of least impact area and the blind watermarking technique. Watermark is embedded within the High-High (HH) sub band. During embedding, watermarked image is dispersed within the band using a pseudo random sequence and a Session key. Watermarked image is extracted using the session key and the size of the image. In this approach the generated watermarked image having an acceptable level of imperceptibility and distortion is compared to the Original retinal image based on Peak Signal to Noise Ratio (PSNR) and correlation value.

## I. INTRODUCTION

Doctors and Medical practitioners often exchange medical images in various diagnostic centres for mutual availability of diagnostic and therapeutic case studies. The communication of medical data through images requires authentication and security. Embedding of watermarks in Digital images can cause distortion in the images. As the images convey information required for detection of diseases, hence any kind of distortion can result in erroneous diagnosis [1, 2, 3, 4]. Embedding of watermark in region of interest (ROI) causes compromise with the diagnosis value of medical image. To achieve medical watermarking technique, proper selection of Non region of interest selection is a crucial task.

Non region of interest is selected between optical disc and macula, at the location of less sensitive diagnostic parametric value with respect to neighboring region within ROI.

Watermarking is added "ownership" information in multimedia contents to prove the authenticity, to verify image integrity, or to achieve control over the copy process [5, 6, 7].

Blind watermarking scheme does not require the original image or any other data. Watermark insertion is done by using an embedding algorithm and a secret key.

Watermarking schemes can be classified either as Spatial Domain or Transformed Domain. Least Significant bit (LSB) [8] insertion is a very simple and common approach to embedding information in an image in special domain. The limitation of this approach is vulnerable to every slight image manipulation. Converting image from one format to another format and back could destroy information hidden in LSBs. Watermarked image can be easily detected by statistical analysis like histogram analysis. This technique involves replacing N number of least significant bit of each pixel of a container image with the data of a watermark. Watermark gets destroyed as the value of N increases. In frequency domain analysis data can be kept secret by using Discrete Cosine Transformation (DCT) [9, 10]. Main limitation of this approach is blocking artifact. DCT pixels are grouped into 8x8 blocks, and transforming the pixel blocks are transformed into 64 DCT coefficients for each block. A modification of a single DCT co-efficient will affect all 64 image pixels in that block. One of the modern techniques of watermarking is Discrete Wavelet Transformation (DWT) approach [11, 12]. In this approach the imperceptibility and distortion of the watermarked image is acceptable. Proposed method of NROI selection is based on Harris Corner Detection, and blind watermarking on retinal image is based on DWT and Spread Spectrum.

## II. METHODOLOGY

### A. Discrete Wavelet Transformation

The wavelet transform describes a multi-resolution decomposition process in terms of expansion of an image onto a set of wavelet basis functions. Discrete Wavelet Transformation has its own excellent space frequency localization property. Application of DWT in 2D images corresponds to 2D filter image processing in each dimension. The input image is divided into 4 non-overlapping multi-resolution sub-bands by the filters, namely LL1 (Approximation coefficients), LH1 (vertical details), HL1 (horizontal details) and HH1 (diagonal details). The sub-band (LL1) is processed further to obtain the next coarser scale of wavelet coefficients, until some final scale "N" is reached. When "N" is reached, 3N+1 sub-bands are obtained consisting of the multi-resolution sub-bands. Which are LLX and LHX, HLX and HHX where "X" ranges from 1 until "N". Generally most of the Image energy is stored in the LLX sub-bands.

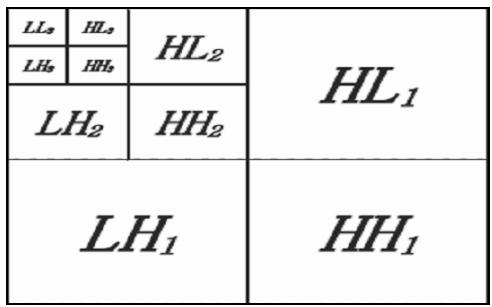

Fig 1. Three phase decomposition using DWT.

The Haar wavelet is the simplest possible wavelet. Haar wavelet is not continuous, and therefore not differentiable. This property can, however, be an advantage for the analysis of signals with sudden transitions.

### B. Code division multiple access (CDMA) spread-spectrum technique

Spread-spectrum technique can be described as a method in which a signal generated in a particular bandwidth when deliberately spread in the frequency domain, results in a signal with a wider bandwidth. If distortion is introduced in this signal by some process such as noise and de-noising, when only certain bands of frequencies maybe damaged, although the message is still in a recoverable state. In spread spectrum communications, the signal energy inserted into any one frequency is too undersized to create a visible artifact and the secret image is scattered over a wide range of frequencies, so that it becomes robust against many common signal distortions. Because of its good correlation properties, noise like characteristics, are easier to generate and resistant to interference, Pseudo Noise Sequences are used for Watermarking.

### C. Harris Corner Detection

Harris corner detector [13,14] is based on the local auto-correlation function of a signal which measuring the local changes of the signal with patches shifted by a small amount in different directions. Given a shift ($\Delta x$, $\Delta y$) to a point (x,y) the auto-correlation function is defined as:

$$c(x,y) = \sum w[I(x_i y_i) - I(x_i + \Delta x, y_i + \Delta y)]^2 \quad \ldots\ldots (1)$$

Where $I(x_i, y_i)$ represents the image function for $(x_i, y_i)$ points in the window W centered around (x, y). Here W is

The Gaussian window is defined as $e^{\frac{-(x+y)^2}{2\sigma^2}}$, where σ defines the width of the window. The shifted image is approximated by a Taylor expansion truncated to the first order terms:

$$I(x_i+\Delta x, y_i+\Delta y) \approx [I(x_i y_i) + [I_x(x_i y_i) I_y(x_i y_i)]] [\Delta x \; \Delta y] \quad \ldots\ldots (2)$$

Where $I_x(x_i, y_i)$ and $I_y(x_i, y_i)$ indicate the partial derivatives with respect to $x_i$ and $y_i$ respectively. With a filter like [-1, 0, 1] and [-1, 0, 1] T, the partial derivates can be calculated from the image by substituting Eqn. (2) in Eqn. (1).

$$c(x,y) = [\Delta x \; \Delta y] \begin{bmatrix} \sum_W (I_x(x_i,y_i))^2 & \sum_W I_x(x_i,y_i) I_y(x_i,y_i) \\ \sum_W I_x(x_i,y_i) I_y(x_i,y_i) & \sum_W (I_y(x_i,y_i))^2 \end{bmatrix} \begin{bmatrix} \Delta x \\ \Delta y \end{bmatrix} = [\Delta x \; \Delta y] C(x,y) \begin{bmatrix} \Delta x \\ \Delta y \end{bmatrix}$$

C(x, y) the auto-correlation matrix captures the intensity structure of the local neighborhood.

For α1 and α2 be Eigen values of C(x, y), three cases may be considered as:

1. Both Eigen values are small signifying uniform region (constant intensity).
2. Both Eigen values are high signifying Interest point (corner)
3. One Eigen value is high signifying contour (edge)

To find out the points of interest, Characterize corner response H(x, y) by Eigen values of C(x, y).

- C(x, y) is symmetric and positive definite that is α1 and α2 are >0
- α1 α2 = det (C(x, y)) = AC – B$^2$

    α1 + α2 = trace(C(x, y)) = A + C

- Harris suggested: the corner response

$$H_{cornerResponse} = \alpha1\ \alpha2 - 0.04(\alpha1 + \alpha2)^2$$

Finally, it is needed to find out corner points as local maxima of the corner response.

## III. PROPOSED ALGORITHM

Retinal Fundus image is collected from The Hamilton Eye Institute Macular Edema Dataset (HEI-MED) (formerly DMED) [16].

### ROI Selection based on the position of the Macula with respect to the Optical Disk:

Step 1. Colored retinal fundus image is converted into gray image from the green channel.

Step 2. Pre-process the gray image to extract retinal blood vessel tree.

Step 3. The image is binarized and Binary area open is applied for removing the small objects.

Step 4. Sobel Edge detection followed by image Thinning is applied on binarized image.

Step 5. Harris Corner Detection Algorithm is applied.

Step 6. Based on the Harris points the best-fit ellipse center is determined (in the Least Squares sense).

Step 7. This center point gives the position of the Macula with respect to the Optical Disk. ROI selection is done based on the presence of Macula.

### Optical Disk Detection:

Step 1. ROI containing optical disk and macula is selected from Retinal Fundus Image and converted into Gray scale image from the green channel of the RGB image for the detection of Optical Disk.

Step 2. The Gray scale image is filtered and normalized.

Step 3. The normalized image is converted to binary based on threshold value. Based on the threshold value selection Optical Disk is selected.

Step 4. Sobel edge detection is applied on binarized image.

Step 5. Harris Corner Detection Algorithm is applied.

Step 6. Maximum Harris Diameter is calculated.

Step 7. The center (O) and diameter (D) is computed and a circle is drawn.

Step 8. The circle (optical disc) is overlain on the original image.

### Macula and NROI Detection:

Step 1. Minimum average intensity points from the overlain original image are detected.

Step 2. At a distance of 1.5D from the centre of the detected optical disc, a rectangular area (search space) of length 1D and height of 0.5D is drawn parallel to X-axis. This search space contains both ROI and NROI, where the NROI is comparatively less required portion conveying any medical information.

Step 3. The distance between center of the optical disk (o) and all average minimum intensity points are computed.

Step 4. Considering only those minimum average intensity points within the search space, the minimum/ maximum and average minimum average intensity points are detected (ROI). These are the nearby points of Macula.

Step 5. The NROI is a square area cropped from the search space with side length of 0.5D where there is no chance of the presence of the macula.

Step 6. Cropped NROI is the area for the image hiding process (Base Image).

### Watermark Embedding:

Watermark embedding process is explained in Fig 2.

Step 1. Base image is converted into Gray scale image from the green channel of the RGB image and decomposed into four sub bands (LL, LH, HL and HH) using DWT.

Step 2. Watermarked image is taken and converted into 1D Vector.

Step 3. Pseudo random 2D sequence is generated by the session based key.

Step 4. HH sub band of the base image is modified by using PN sequence depending upon the content of the secret 1D image vector to be embedded.

Step 5. Four sub bands including modified sub band are combined to generate the Watermarked image using IDWT.

Step 6. Red and Blue channel is concatenated with watermarked image to get the color watermarked base image.

Step 7. Color watermarked image is overlain on the selected NROI area of original Retinal Fundus image.

Step 8. Finally, the color Retinal Fundus image is reconstructed.

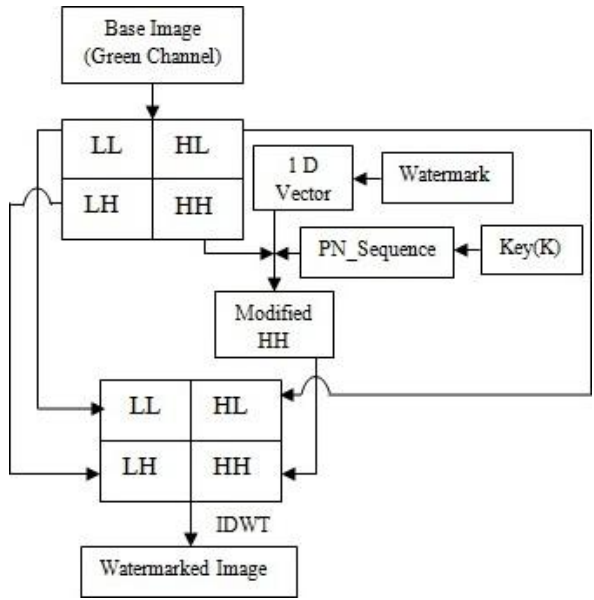

Fig 2: Watermark Embedding Process

**Watermark Extraction:**

Watermark extraction process is explained in Fig 3.

NROI is selected in the same way as stated earlier from the color watermarked Retinal Fundus image using Harris corner detection. NROI is converted into gray scale image from the green channel of the RGB image.

Step 1. Session key and Sizes of the watermark image is sent to the intended receiver via a secret communication channel.

Step 2  Watermark image can be recovered from the watermarked image using correlation function and knowing the size of the original image.

Step 3. Extracted watermark image is filtered to remove the unwanted signal.

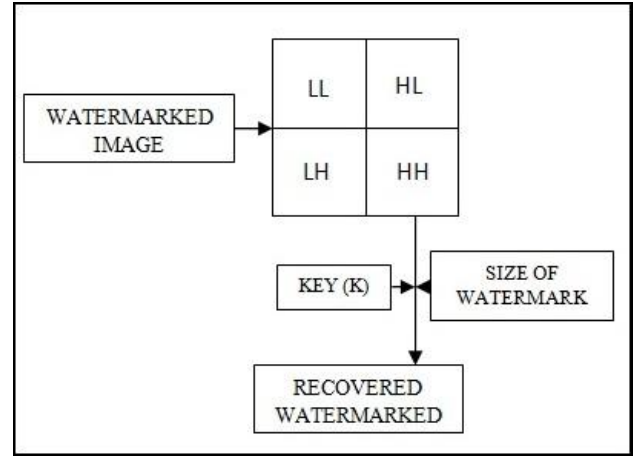

Fig 3: Watermark Extraction Process

## IV. EXPLANATION OF THE ALGORITHM

**Detection of the position of the Macula with respect to the Optical Disk:**

The colored Retinal Fundus image is taken as an input image. The green channel exhibits the best contrast between foreground and the background in the RGB images, whereas the red and blue channels tend to be noisier. Naturally gray image from the green channel is taken into consideration for present study. The approximate background is estimated using a 7x7 Wiener filter and a 22x22 median filter on the gray image. In the pre-processing steps, extraction of the filtered image from the original filtered gray image leads to a brighter area than the background. Generated image is converted to binary data based on the threshold value. Sobel edge detection followed by image thinning is applied on binary image to obtain the retinal blood vessel tree. Harris Corners are applied on the edge detected image. Fitting an ellipse on the obtained Harris points in the plane is done by using Least-Squares Fitting [15]. The sum of the squares of the distances to the given Harris points is minimal for the ellipses, and the concerned matching is referred to as "best-fit". The centre of the best-fit ellipse is computed. This point gives the position of the Macula with respect to the Optical Disk.ROI selection is done based on the presence of Macula in the retinal image.

**Optical Cup and Macula Detection procedure:**

The Retinal Fundus image is taken as an input image. The ROI is cropped from the image. The green channel exhibits the best contrast between foreground and the background in the RGB images, whereas the red and blue channels tend to be noisier. In the cases of OD detection green channel is taken into consideration for the conversion of RGB to gray. The OD area is getting darker in the gray image. The approximate background is estimated using a 7x7 Wiener filter and a 22x22 median filter on the gray image.

Complement operation is performed on the filtered image followed by normalization by extracting the complemented image from the original gray image. These lead to a brighter area than the background. The normalized image binary data based on the threshold value is converted. Sobel edge detection followed by Harris Corner detection algorithm is applied on the edge detected binarized image. Maximum Harris Diameter (HD) is calculated considering all the Harris points. Center point (o) of the HD and radius(R) is computed and finally a circle is drawn based on the R and o value. Circle is overlain on the original image.

Minimum average intensity points from the overlain original image are detected. A search space contains both ROI an NROI is cropped based on the proposed method.

**Watermark Embedding procedure:**

The NROI is converted into Gray scale image from the green channel. Using DWT, the gray NROI image is decomposed into four sub bands (LL, LH, HL and HH).

A binary image (watermark) is taken and converted into 2 one dimensional vectors. Pseudo random sequence is generated using a session based key and the size of HH sub band of the gray image. Each of the bits of the binary is embedded in HH sub-band depending upon the elements of the one dimensional vector and the pseudo random sequence. The general equation used to embed the secret image is:

$$IS(x, y) = I(x, y) + k \times S(x, y) \quad \ldots\ldots\ldots (3)$$

In which $I(x, y)$ representing the selected DWT sub band of the gray image, $IS(x, y)$) is the modified gray image, K denotes the amplification factor that is usually used to adjust the invisibility of the secret image in corresponding sub band. $S(x, y)$ is the pseudo random sequence.

Taking all the sub bands including the modified HH sub-band, watermarked image is obtained applying IDWT (Inverse Discrete Wavelet Transformation). Red and Blue channel is concatenated with gray watermarked image to get the color watermarked base image and overlain on the selected NROI area of original Retinal Fundus image to get back the watermarked color Retinal Fundus image.

**Watermark Extraction procedure:**

NROI is selected from the color watermarked Retinal Fundus image using Harris corner detection as stated in the proposed method. NROI is converted into gray scale image from the green channel of the RGB image.

The session key and the size of the HH sub-bands of watermarked image is provided to the intended receiver through a secrete communication channel.

Select the HH sub-band of watermarked image after applying DWT. The pseudo random sequence (PN) is regenerated using the same session based key which was used in the secret image embedding procedure. The correlation between the selected watermarked sub-band and the generated pseudo random sequence is calculated. Each correlation value is compared with the mean correlation value. If the calculated value is greater than twice of the mean, then the extracted watermark bit is taken as a 0, otherwise 1. The recovery process then iterates through the entire PN sequence until all the bits of the watermark image have been recovered.

Filter is used on recovered secrete images to remove unwanted signals.

### V. RESULTS AND DISCUSSION

MATLAB 7.0.1 Software is extensively used for the study of the retinal watermarking embedding and extraction process. Concerned images obtained in the result are shown in Fig. 4 through 23.

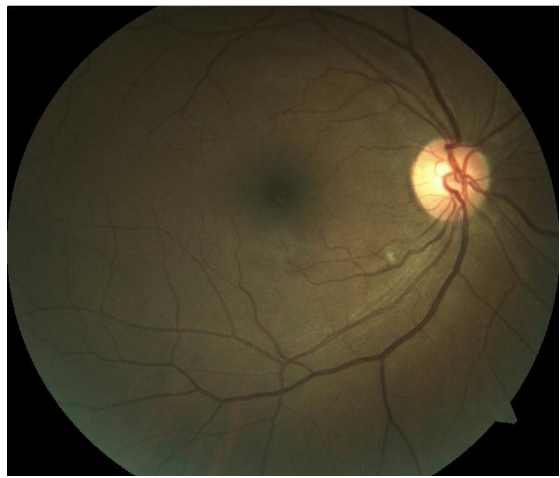

Fig. 4: Retinal Fundus Image

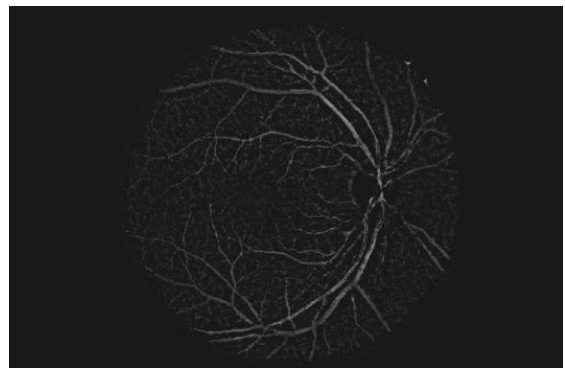

Fig 5. Pre-processed Gray Image

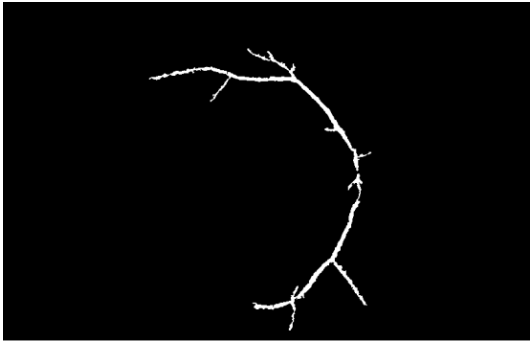

Fig 6. Extracted Retinal Blood Vessel Tree

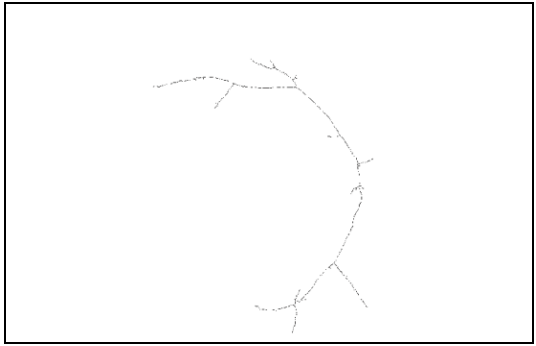

Fig 7. Vessel Tree after Thinning

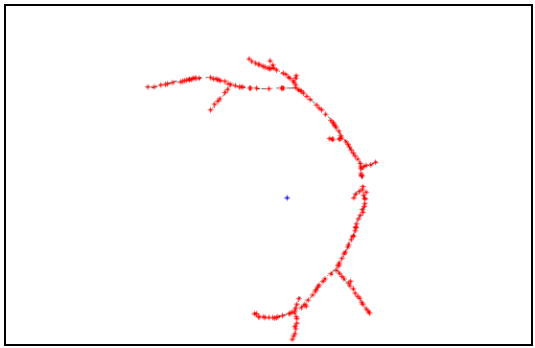

Fig 8. Harris Corner Points and Best-fit ellipse   center.

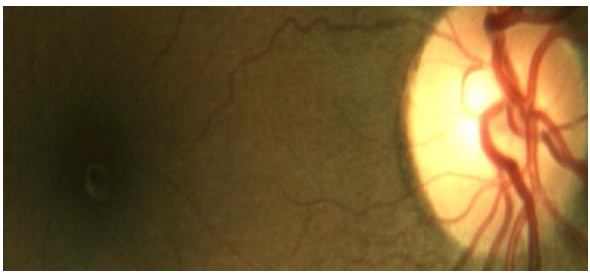

Fig 9. Cropped ROI containing Optical Disk and Macula from the Original Image to the Direction of Best-fit ellipse center.

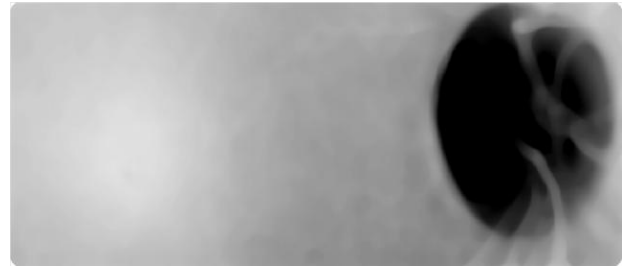

Fig. 10. Green Channel of ROI

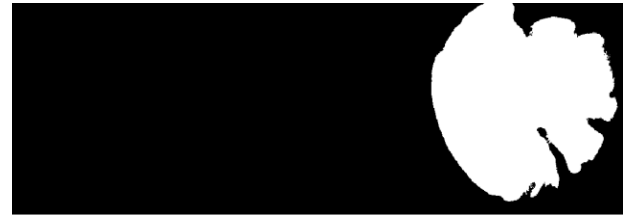

Fig 11. Filtered and normalized Image followed by Binarization.

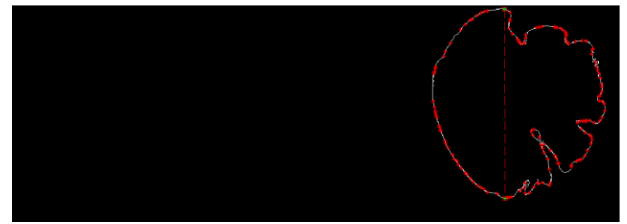

Fig 12. Maximum Harris Diameter in the Edge Detected Image.

Max Harris Point (854,7) and (752,80) Harris Dia=325.0015 Harris Radius=162.5008 Optical Disk Center (854.5000, 169.5000)

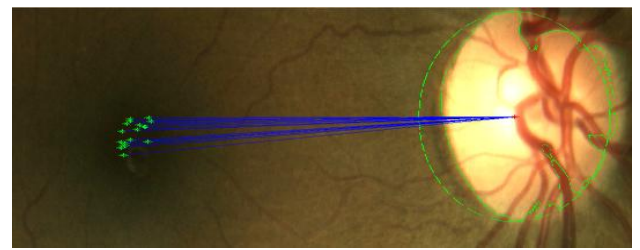

Fig. 13. Minimum Average Intensity Points in ROI

Number of minimum average intensity points in ROI is 17. Distance between all 17 points with Optical Disk Center is as follows.
670.6642,671.2544,667.8791,667.1585,664.9906,663.7985, 659.5836,653.5323,654.4635,652.5094,640.7968,637.6225, 630.6905,628.6911,624.6875,622.5018,618.5455

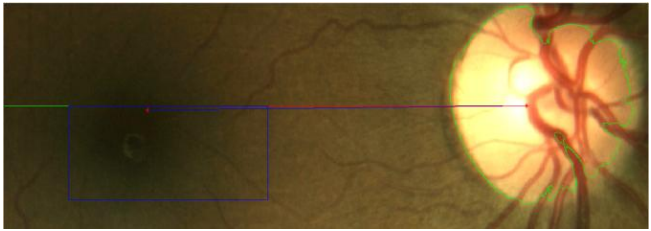

Fig. 14: Search Space

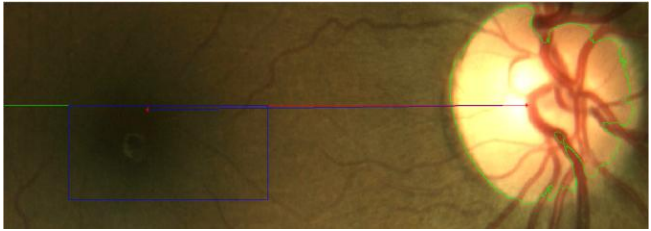

Fig. 15: Minimum Average Intensity point among all the Minimum Average Intensity Points in the Search Space.

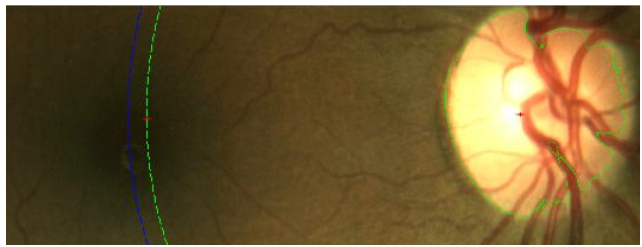

Fig. 16: Minimum Average Intensity point among all the of Minimum Average Intensity Points in the Search Space. Among all 17 points, 9 points are there within the search space.

Minimum of all Minimum Average Intensity point distances within the search area is 618.5455. Average of all Minimum Average Intensity point distances within the search space is 648.7865. Maximum of all Minimum Average Intensity point distances within the search space is 671.2544

Position of the minimum average intensity point distance within the search area is (106.9965, 431.9980)

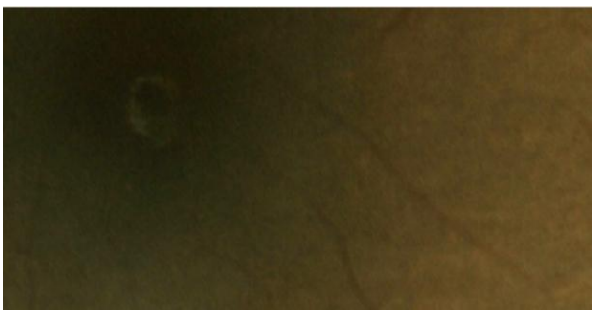

Fig. 17: Search Area (ROI+NROI) Height of Search space is 0.5D and length is 1.5D.

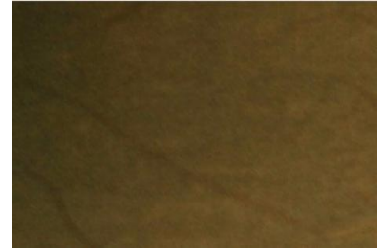

Fig. 18: NROI Height of NROI is 0.5D and length is 0.5D.

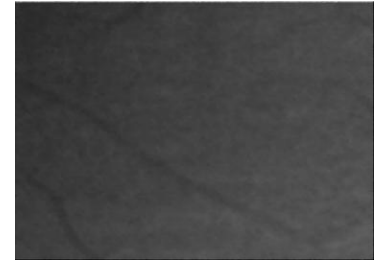

Fig. 19: Green Channel of NROI

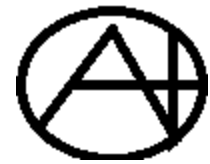

Fig. 20: Watermark

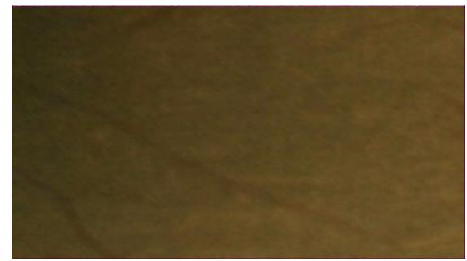

Fig. 21: Watermarked Image

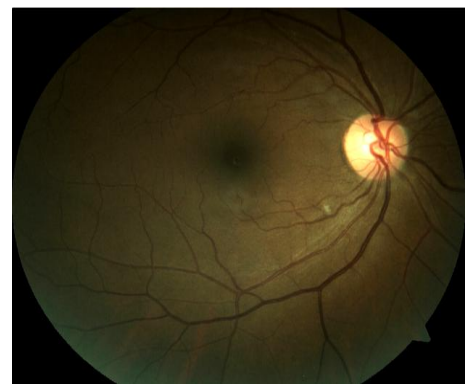

Fig. 22: Overlain Watermarked Image on Original Image (Final Watermarked Image).

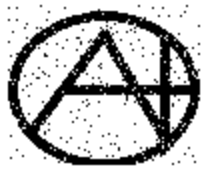

Fig. 23: Recovered Watermark

**Peak Signal to Noise Ratio (PSNR)**

It measures the quality of a watermarked image. This is basically a performance metric which uses to determine perceptual transparency of the watermarked image with respect to original image:

$$PSRN = \frac{MN\,max_{x,y} P_{x,y}^2}{\sum_{x,y}(P_{x,y} - \bar{P}_{x,y})^2} \quad (1)$$

Where, M and N are number of rows and columns in the input image,

$P_{x,y}$ is the original image and

$\bar{P}_{x,y}$ is the watermarked Image.

PSNR between gray NROI Image and watermarked Image is 29.7910 shown in Table1.

Table. 1

|  | PSNR |
|---|---|
| Gray NROI Image vs. Watermarked Image | **29.7910** |

**Correlation coefficient**

After secret image embedding process, the similarity of original image x and watermarked image x' is measured by the standard correlation coefficient as follows:

$$\text{Correlation} = \frac{\sum(x-x')(y-y')}{\sqrt{(x-x')^2}\sqrt{(y-x')^2}} \quad (2)$$

Where y and y' are the discrete wavelet transforms of x and x'.

Correlation between the watermark and recovered watermark after applying filter is 0.8894 shown in the Table2.

Table. 2

| Correlation between original watermark image and recovered watermark image | Image1 |
|---|---|
|  | **0.8894** |

## VI. CONCLUSION

Proposed technique is useful in telemedicine applications for authentication of the source of the information. In this present work as Watermark is embedded in the HH sub band of the original image, there is a small visual change in between the original image and the watermarked image generating some imperceptibility of image data. But due to strong security aspects this small amount of imperceptibility is acceptable. As the watermark image is embedded in the NROI the diagnosis value of medical image is maintained within the tolerance level. The values of correlation and PSNR are very much encouraging regarding the faithfulness of the reconstruction of the image.